\documentclass[10pt,twocolumn,letterpaper]{article}

\usepackage{ijcb}
\usepackage{times}
\usepackage{epsfig}
\usepackage{graphicx}
\usepackage{amsmath}
\usepackage{amssymb}

\usepackage{makecell}
\usepackage{booktabs}
\usepackage{subfig}
\usepackage{multirow}

\ijcbfinalcopy 

\ifijcbfinal\fi
\begin{document}

\title{On the (Limited) Generalization of MasterFace Attacks and \\ Its Relation to the Capacity of Face Representations}


\newcommand\Mark[1]{\textsuperscript#1}

\author{Philipp Terh\"{o}rst\Mark{1}\Mark{2}, Florian Bierbaum\Mark{3}, Marco Huber\Mark{2}\Mark{3}, Naser Damer\Mark{2},\\ Florian Kirchbuchner\Mark{2}\Mark{3}, Kiran Raja\Mark{1}, Arjan Kuijper\Mark{2}\Mark{3}\\
\Mark{1}Norwegian University of Science and Technology, Gjøvik, Norway\\
\Mark{2}Fraunhofer Institute for Computer Graphics Research IGD, Darmstadt, Germany\\
\Mark{3}Technical University of Darmstadt, Darmstadt, Germany\\
}

\maketitle
\thispagestyle{empty}

\begin{abstract}
\vspace{-1mm}
A MasterFace is a face image that can successfully match against a large portion of the population.
Since their generation does not require access to the information of the enrolled subjects, MasterFace attacks represent a potential security risk for widely-used face recognition systems.
Previous works proposed methods for generating such images and demonstrated that these attacks can strongly compromise face recognition. 
However, previous works followed evaluation settings consisting of older recognition models, limited cross-dataset and cross-model evaluations, and the use of low-scale testing data.
This makes it hard to state the generalizability of these attacks.
In this work, we comprehensively analyse the generalizability of MasterFace attacks in empirical and theoretical investigations.
The empirical investigations include the use of six state-of-the-art FR models, cross-dataset and cross-model evaluation protocols, and utilizing testing datasets of significantly higher size and variance.
The results indicate a low generalizability when MasterFaces are training on a different face recognition model than the one used for testing.
In these cases, the attack performance is similar to zero-effort imposter attacks.
In the theoretical investigations, we define and estimate the face capacity and the maximum MasterFace coverage under the assumption that identities in the face space are well separated. 
The current trend of increasing the fairness and generalizability in face recognition indicates that the vulnerability of future systems might further decrease.
Future works might analyse the utility of MasterFaces for understanding and enhancing the robustness of face recognition models.

%
%
%
%
%

\end{abstract}



\vspace{-5mm}
\section{Introduction}


In face recognition (FR), a MasterFace is a face image that can be successfully matched  against a large portion of the population \cite{DBLP:conf/fgr/ShmelkinWF21}.
Such a face can be used to impersonate any user without having access to information about the enrolled subject \cite{DBLP:conf/icb/NguyenYEM20}.
Since face recognition systems (FRS) are spreading worldwide and are increasingly involved in our daily life activities \cite{DBLP:journals/ijon/WangD21a}, successful MasterFace attacks pose a great threat to these systems.

\begin{figure}
\includegraphics[width=0.45\textwidth]{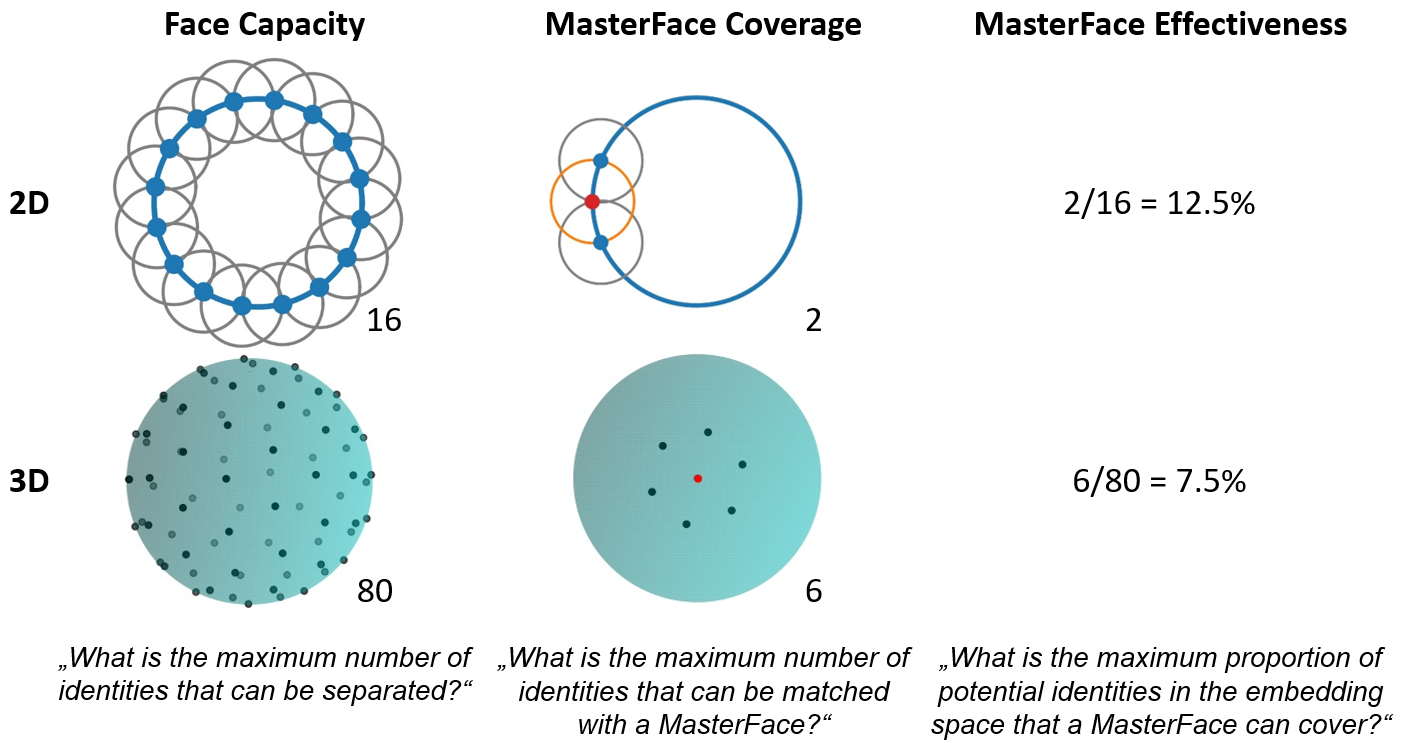}
\caption{Visualization of theoretical concepts - Assuming an embedding space (on a unit-sphere) with perfect identity-separation, the three concepts are visualized for two and three-dimensional embeddings for a fixed decision threshold $r=0.4$. In 2D, an optimal MasterFace can cover 12.5\% of the identities. In 3D, this drops to 7.5\% indicating that for higher-dimensional representations the effectiveness might further decrease.
\vspace{-3mm}
}
\label{fig:Visualization}
\end{figure}

Consequently, MasterFaces attracted attention in both media and academia.
For instance, the work from Shmelkin et al. \cite{DBLP:conf/fgr/ShmelkinWF21} won the Google Best Paper Award at IEEE International Conference on Automatic Face and Gesture Recognition 2021 for their work on generating MasterFaces.
In that work, the authors report that a single face can cover more than 20\% of the identities in the used database (for a given FR model at a threshold for a false match rate of $10^{-3}$).
Also, Nguyen et al. \cite{DBLP:conf/icb/NguyenYEM20, DBLP:journals/corr/abs-2109-03398} proposed a similar method for generating MasterFaces and demonstrated that these attacks can successfully compromise FRS.
However, these works (a) make use of older FRS, (b) consist of only limited cross-dataset and cross-model evaluations, and (c) conduct their experiments on testing data of small size and variance.
This does not allow to make significant statements on the generalizability of these attacks.

In this work, we empirically and theoretically analysed the generalizability of MasterFace attacks.
Contrarily to previous works, our experiments (a) make use of six state-of-the-art FR models, (b) are based on cross-dataset and cross-model evaluations, and (c) utilize testing datasets that are of a significantly higher size and variance.
To ensure comparability with previous works \cite{DBLP:conf/fgr/ShmelkinWF21, DBLP:conf/icb/NguyenYEM20}, we use the original MasterFace images reported in the Arxiv versions of these papers (CC-BY-4.0) for the experiments.
The empirical investigations showed that only an insignificant proportion of the testing identities are successfully matched against the MasterFaces and that the comparison scores resulting from comparisons with MasterFaces closely resembles the score distribution of zero-effort imposter.
Consequently, in our experiments, no utilized FRS showed any significant vulnerability to MasterFaces attacks.
The observations lead to the assumption that successful MasterFace attacks exploit regions in the embedding space of a specific FRS characterized by poor identity-separability.

To assess the threat of MasterFace attacks regarding the rapid development of more generalizable FRS, we conduct a theoretical analysis on the maximum effectiveness of MasterFace attacks assuming an FRS that can perfectly differentiate between identities.
For this, we define and estimate the face capacity as the maximum number of identities that can be differentiated and the maximum MasterFace Coverage as the maximum number of perfectly-separable identity representations that are successfully matched with a MasterFace template.
We define the ratio of the coverage and the capacity as the maximum MasterFace effectiveness.
The theoretical investigations showed that the face capacity is several magnitudes higher than the maximum MasterFace coverage demonstrating that the maximum MasterFace effectiveness becomes insignificantly low when dealing when a perfect identity-separation is achieved.
In Figure \ref{fig:Visualization}, this is visualized for lower-dimensional representations.
With the current trend of increasing the generalizability of FRS, this indicates that the vulnerability of MasterFace attacks might further decrease in the future.
In summary, the empirically and theoretically investigations showed that MasterFace attacks do not generalize well to unseen data and FR models.

%
%
%
%
%
%
%
%
%
%
%
%
%
%
%
%
%
%

\section{Background}

\subsection{The Dodding Zoo and MasterFaces}

Generally, biometric systems perform differently on different groups of users \cite{DBLP:journals/corr/abs-2103-01592, 8636231, FRVT2019}.
In \cite{DBLP:journals/pami/YagerD10, DBLP:conf/interspeech/DoddingtonLMPR98}, several user groups have been characterized depending on their effect on the biometric system.
This concept, formalized by Doddington et al. \cite{DBLP:conf/interspeech/DoddingtonLMPR98}, is known as biometric menagerie or Doddington Zoo.
Originally, this consists of four members: Sheep, Goats, Lambs, and Wolves.
Sheep match well against themselves and poorly against others.
Goats are generally difficult to match.
Lambs are vulnerable to impersonation attacks and Wolves are highly successful in impersonating lambs.
MasterFace attacks aim at creating synthetic and highly-effective wolve-samples \cite{DBLP:conf/icb/NguyenYEM20}.
They aim for generating face images that pass biometric identity verification for a large portion of the population and thus, can be used to impersonate, with a high probability of success, any user, without having access to any user-information \cite{DBLP:conf/fgr/ShmelkinWF21}.
This is the reason why in \cite{DBLP:conf/icb/NguyenYEM20}, MasterFace attacks are also named Wolf attacks.

\subsection{Generation of MasterFaces}

Previous works \cite{DBLP:conf/fgr/ShmelkinWF21, DBLP:conf/icb/NguyenYEM20, DBLP:journals/corr/abs-2109-03398} share a similar idea on the MasterFace generation process.
The works make use of a pre-trained StyleGAN \cite{DBLP:conf/cvpr/KarrasLA19} model $\mathcal{G}$ for the generation of a high-quality face image $\mathcal{I}$ from a given latent vector $x$ and search in the latent space for such a MasterFace vector  
\begin{align}
x_{MF} = \arg\max_x \sum_{\mathcal{I}\in \mathcal{D}} f(\mathcal{M} (\mathcal{G}(x)), \mathcal{M}(\mathcal{I}), \theta)
\end{align}
that maximizes its coverage with the images $\{I\}$ from a given dataset $\mathcal{D}$.
Here, $\mathcal{M}(\mathcal{I})$ is a face recognition model that takes an image $\mathcal{I}$ as an input and output a face embedding $z$. 
The function $f(x,y,\theta)$ is a matching function that compares the similarity between the embeddings $x$ and $y$ and returns 1 or 0 depending on a decision threshold $\theta$.
To solve this optimization problem, a Covariance Matrix Adaptation Evolution Strategy (CMA-ES) \cite{DBLP:conf/gecco/PitraBRH17a} is used.
In \cite{DBLP:conf/fgr/ShmelkinWF21}, multiple evolutionary strategies are compared and a novel approach is proposed that uses a neural network to guide the search without a fitness evaluation.
The final MasterFace image $\mathcal{I}_{MF}$ is given by $\mathcal{G}(x_{MF})$.

%

\subsection{Limitations}

The idea of creating a synthetic biometric sample to match a larger portion of the population was first introduced for fingerprint images known as MasterPrints \cite{DBLP:journals/tifs/RoyMR17, DBLP:conf/icb/RoyMTR18, DBLP:conf/btas/BontragerRTMR18}.
MasterPrints build on the fact that in some applications, such as smartphones, the
fingerprint sensor is small and thus, only partial fingerprint images are used for matching \cite{DBLP:conf/btas/BontragerRTMR18}.
In contrast to MasterPrints, works on MasterFaces report high performances even when non-partial information is available.
However, this might be explained by the limitation of their experimental setups.

\begin{itemize}
\item \textbf{Weak assumption} - Previous works \cite{DBLP:conf/fgr/ShmelkinWF21, DBLP:conf/icb/NguyenYEM20, DBLP:journals/corr/abs-2109-03398} build on the assumption that many identities might cluster at specific points in the embedding space to motivate the existence of MasterFace images. 
However, this means that the identities of such a cluster are already similar to each other and thus, might already match each other.
This is supported by the observation that generated MasterFaces mainly show very old or young faces \cite{DBLP:conf/fgr/ShmelkinWF21}.
Since, current FRS aim to maximize the margin between identities in the embedding space, and thus to distribute in the whole space, in this work, we will theoretically analyse how well these attacks work if this assumption is not given.
Moreover, building on the assumption of clustered identities in the embedding space makes the attacks strongly dependent on the face recognition model and thus, shows the need for investigating the inter-model generalizability.


\item \textbf{Older face recognition models} - Previous works limited their analysis to the usage of rather old face recognition models that have less advanced generalization mechanisms. 
For instance, in \cite{DBLP:conf/fgr/ShmelkinWF21}, Dlib \cite{DBLP:journals/jmlr/King09} (2009), SphereFace \cite{DBLP:conf/cvpr/LiuWYLRS17} (2017), and FaceNet \cite{DBLP:conf/cvpr/SchroffKP15} (2015) were investigated.
In this work, we will analyse the effectiveness of MasterFace attacks on more recent and advanced FR models building on stronger generalization techniques.


\item \textbf{Limited cross-dataset/model evaluations} - In previous works \cite{DBLP:conf/fgr/ShmelkinWF21, DBLP:conf/icb/NguyenYEM20, DBLP:journals/corr/abs-2109-03398}, the MasterFace generation model was mainly (a) optimized for the effectiveness of a specific FR model used for testing and (b) evaluated with data coming from the same distribution as the training data\footnote{Please note that no cross-model generalizability was claimed in \cite{DBLP:conf/fgr/ShmelkinWF21}.}.
Both points might easily result in overfitted MasterFaces and demonstrate the need for cross-dataset and cross-model evaluations.
In this work, we focus on these two evaluation settings to elaborate on the generalizability of these attacks.

\item \textbf{Small testing data} - Previous works conducted their experiments with low-scale testing data.
This includes the use of the LFW dev set (1711 images) \cite{DBLP:conf/fgr/ShmelkinWF21, DBLP:conf/icb/NguyenYEM20}, 
 and the dev set of the MOBIO dataset (58 identities) \cite{DBLP:conf/icb/NguyenYEM20,DBLP:journals/corr/abs-2109-03398}.
Since the amount and variance of testing data might strongly affect the performance of MasterFace attacks, we will analyse this for both aspects on a larger scale than previous research.



\end{itemize}

%

\section{Methodology}
\label{sec:Methodology}

Besides the comprehensive empirical evaluation, we propose a theoretical analysis of the effectiveness of MasterFace attacks.
Previous attacks \cite{DBLP:conf/fgr/ShmelkinWF21, DBLP:conf/icb/NguyenYEM20, DBLP:journals/corr/abs-2109-03398} build on the assumption that many identities might cluster at specific points in the embedding space.
In the following, we will analyse the effectiveness of optimal MasterFace attacks when such an assumption is not given.
This is achieved in three steps.
First, the capacity of a face space is calculated stating the optimal number of perfectly separable identities in this space.
Second, the maximum coverage of a MasterFace is computed in the same context.
Third, we will extrapolate the face capacity and MasterFace coverage to analyse which portion of the face space can maximally be covered by a MasterFace in high dimensions.


\subsection{The Capacity of the Face Space}

In the following, we assume a face space on a $d$-dimensional sphere (unit-length face embeddings) in which every identity is perfectly separated and matching function $r$ stating a non-match if two face embeddings $x^{(i)}$ and $x^{(j)}$ have an Euclidean distance of $D(x^{(i)}, x^{(j)})>r$. 
The capacity $N$ of a face space (defined over $d$ and $r$) describes the maximum number of identities that can be encoded in such a space with an optimal separation.
For a given $d$ and $r$, the following optimization problem
\begin{align}
\min_x \sum_{i=1}^n \sum_{j=i+1}^n \frac{1}{D(x^{(i)}, x^{(j)})} \\ \text{with} \qquad ||x^{(i)}||=1 \qquad \forall \,\,\, i\in [1,\dots,n]
\end{align}
equally distributes $n$ identities in the face space.
The optimization constraint $||x^{(i)}||=1$ ensures that all embeddings lie on a sphere and that the conclusions also generalize to cosine similarity (since for unit-length embeddings both measures only differ by a constant factor).
By iteratively solving this problem with increasing $n$ and determining the nearest neighbour distance $r'$, the capacity of the face space is determined by the greatest $n$ that fulfils $r'<r$.

\subsection{Maximum MasterFace Coverage}

The maximum MasterFace coverage $\overline{N}$ describes the maximum number of perfectly-separable identity embeddings $x^{(i)}$ that are matched with a MasterFace embedding $x^{(MF)}$.
To achieve this, for a given $d$ and $r$, the optimization problem
\begin{align}
\min_x \frac{1}{n} \sum_{i=1}^n \left( D(x^{(i)}, x^{(MF)}) - r  \right)^2 \label{eq:Coverage}\\
\text{with} \,\,\, D(x^{(i)}, x^{(j)}) > r \,\,\, \text{and} \,\,\, ||x^{(i)}||=1 \,\,\, \forall \, i,j
\end{align}
can check if $n$ identity embeddings $x^{(i)}$ can be placed around the MasterFace embedding $x^{(MF)}$ such that these can be successfully matched against the MasterFace embedding without being matched against each other.
Minimizing Eq. \ref{eq:Coverage} ensures that the MasterFace can successfully match the $n$ identities while the optimization constraint $D(x^{(i)}, x^{(j)}) > r$ ensures that these identities are still separable from each other.
The highest $n$ that fulfils the described criteria determines the maximum MasterFace coverage $\overline{N}$.

\subsection{Maximum MasterFace Effectiveness}
\label{sec:MaximumMFEffectiveness}

The maximum MasterFace effectiveness describes the maximal proportion of all perfectly-separable identities in the embedding space that a MasterFace can successfully match with.
Consequently, we can define the maximum MasterFace effectiveness $\eta(r,d)=\frac{\overline{N}(r,d)}{N(r,d)}$ over the ratio of the MasterFace coverage $\overline{N}$ and the face capacity $N$.
Since the calculations of these values require computationally-expensive efforts that are not feasible for higher dimensions, we solve these explicitly for $d\in[3, 10]$ and $r\in[0.65, 1.35]$ with a Sequential Least Squares Programming (SLSQP) algorithm and extrapolate these to get an estimate about the magnitude of $\eta(r,d)$ for larger $d$.
We found that the face capacity can be accurately described through
\begin{align} 
N(r,d) = \exp \left[ d \, (\alpha + \gamma \, r) + \beta + \delta \, r   \right]
\label{eq:FitCapacity}
\end{align}
 while the extrapolation of the MasterFace coverage follows
\begin{align}
\overline{N}(r,d) = (\overline{\alpha} \, d^3 + \overline{\beta}) \cdot \sigma\left( \phi \, (   r^{\epsilon}-1) \right) + \overline{\gamma} \, d^3 + \overline{\delta}.
\label{eq:FitCoverage}
\end{align}
Here, $\exp(\cdot)$ and $\sigma(\cdot)$ are the exponential and the sigmoid functions.
In Eq. \ref{eq:FitCoverage}, we manually set $\phi=10000$ and $\epsilon=0.0005$ to determine the shape of the sigmoid function $\sigma(\cdot)$ regarding $r$.
The learned parameters for both functions are shown in Table \ref{tab:LearnedParameters}.
In Section \ref{sec:TheoreticalAnalysis}, we will first discuss the suitability of the capacity and coverage approximations before discussing its implications on the MasterFace effectiveness under the assumption of face space with perfect identity-separation.

%
%
%
%
%
%
%

\begin{table}[]
\renewcommand{\arraystretch}{1.0}
\centering
\caption{Learned parameters for the capacity and coverage.}
\label{tab:LearnedParameters} 
\begin{tabular}{lrrrr}
\Xhline{2\arrayrulewidth}
         & \multicolumn{1}{c}{$\alpha$} & \multicolumn{1}{c}{$\beta$} & \multicolumn{1}{c}{$\gamma$} & \multicolumn{1}{c}{$\delta$} \\
         \hline
Capacity & 0.993                 & 3.701                & -0.436                & -3.706                \\
Coverage & -0.172                & 8.258                & 0.153                 & 0.315                \\
\Xhline{2\arrayrulewidth}
\end{tabular}
\end{table}

\section{Experimental setup}

\subsection{Databases and images}

The experiments were conducted on three publicly-available datasets.
The Adience dataset \cite{Eidinger:2014:AGE:2771306.2772049} consists of 26k images from over 2k different subjects.
The images of the Adience dataset possess a wide range in terms of image quality and additionally consists of many young faces.
Contrarily to previous works that focus only on around 10\% of the images and identities of LFW \cite{LFWTech} for testing, we analyse the performance on the whole database containing 13k images of over 5k identities. 
The ColorFeret database \cite{DBLP:journals/pami/PhillipsMRR00} consists of 14k high-resolution face images from over 1k different individuals.
The data possess a variety of face poses (from frontal to profile) and facial expressions under well-controlled conditions.
For the experiments, the reported MasterFace images from both groups that worked on this problem  \cite{DBLP:conf/fgr/ShmelkinWF21, DBLP:conf/icb/NguyenYEM20} were used as included in their respective Arxiv papers (CC-BY-4.0).
For the sake of simplicity, we refer to the images from \cite{DBLP:conf/icb/NguyenYEM20} as WolfAttacks and from \cite{DBLP:conf/fgr/ShmelkinWF21} as MasterFace in the experiments.

\subsection{Evaluation metrics}

Following the international standard for biometric evaluation \cite{ISO_Metrik}, we report the face verification error in terms of false non-match rate (FNMR) at a threshold for a fixed false match rate (FMR).
Extending the mean set coverage score from \cite{DBLP:conf/fgr/ShmelkinWF21}, we introduce the concept of sample- and identity-coverage to analyse the effectiveness of MasterFace attacks on the image- and identity-level.
Comparing the MasterFaces against every image of a given database, the sample-coverage states the mean portion of images that are successfully matched per MasterFace image.
On the other hand, the identity-coverage states the mean portion of identities that are matched successfully requiring at least one image per identity to be matched with a MasterFace.

\subsection{Utilized models}

In the experiments, six state-of-the-art FR approaches are utilized for evaluating the MasterFace attacks.
This include 
ArcFace \cite{DBLP:conf/cvpr/DengGXZ19} (2019),
CurricularFace \cite{DBLP:conf/cvpr/HuangWT0SLLH20} (2020),
Elastic-Arc and Elastic-Cos \cite{DBLP:journals/corr/abs-2109-09416} (2021),
MagFace \cite{DBLP:conf/cvpr/MengZH021} (2021), and
QMagFace \cite{DBLP:journals/corr/abs-2111-13475} (2021).
For the experiments, we used the (ResNet-100/iResNet-101) models from the official repositories of the authors to extract the face templates.
Comparing two templates was done based on the standard cosine similarity.


%
%
%

\subsection{Investigations}

This work analyses the generalizability of MasterFace attacks through empirical and theoretical investigations.
The empirical investigations aim to analyse the vulnerability of six state-of-the-art FRS to MasterFace attacks.
This includes analysing sample- and identity-coverage rates as well as analysing the results on the score-level.

To approximate the vulnerability of future FRS with increasing identity-separation, the theoretical investigations analyses the face capacity in relation to the maximum MasterFace coverage under the assumption of a face space with perfect identity-separation.
This allows studying the maximum MasterFace effectiveness.

%
%
%
%

\section{Results}

\subsection{Empirical Analysis}

\paragraph{Investigating Coverage-Rates:}
To analyse the vulnerability of six state-of-the-art FRS on MasterFace attacks, Table \ref{tab:EmpiricalResults} shows sample- and identity-coverage rates on three datasets for WolfAttack and MasterFace images on the mentioned FRS.
To cover a wide range of applications, the decision thresholds range between FMR of $10^{-1}$ to $10^{-4}$.

Generally, the sample- and identity coverage drops drastically for lower FMRs and the investigated FRS show no vulnerability to the MasterFace attacks.
This is also true for LFW.
Despite that both methods use large parts of LFW as training data, the coverage rates are close to zero.
Consequently, we assume that the main factor for the reported success of MasterFace attacks rely on the fact that these attacks are optimized for very specific FR models.

 
In \cite{DBLP:conf/icb/NguyenYEM20}, FMRs of 6-35\% (at a threshold for FMRs between 1-3\%) were reported for the WolfAttacks depending on the testing data and the FRS. 
This relates to our sample-coverage at an FMR around $10^{-1}$ which varies between 2.96-8.94\% in our experiments.
In \cite{DBLP:conf/fgr/ShmelkinWF21}, over 40\% of the testing identities were matched with 10 MasterFaces on three FRS.
However, this was achieved by optimizing the MasterFace generation on the target model and the testing data.
For the identity-separated evaluation, \cite{DBLP:conf/fgr/ShmelkinWF21} reported  an identity-coverage between 17-22\% at an FMR threshold of $10^{-3}$.
This is significantly lower than the observed performance of 0.01-0.31\% at the same FMR threshold in our experiments using larger datasets and FR models that the attacks are not trained on.

Overall, the results show a very limited generalizability of the MasterFace attacks on recent FRS.
The experiments demonstrate that the utilized FRS show no significant vulnerability to any MasterFace attacks if these attacks are not directly optimized on the deployed FRS.
This holds especially true for FMR of $10^{-3}$ and smaller as recommended by the European Boarder Guard Agency Frontex \cite{FrontexBestPractice}.
In this range, the identity-coverage lies below 0.31\% in all investigated cases.

\begin{figure*}
\centering
\captionsetup[subfigure]{justification=centering}
\subfloat[Adience\protect\\ArcFace]{%
     \includegraphics[width=0.135\textwidth]{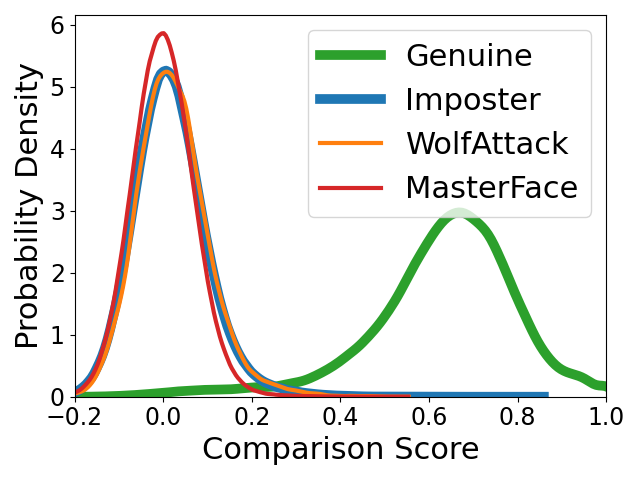}} 
\subfloat[Adience\protect\\CurricularFace]{%
     \includegraphics[width=0.135\textwidth]{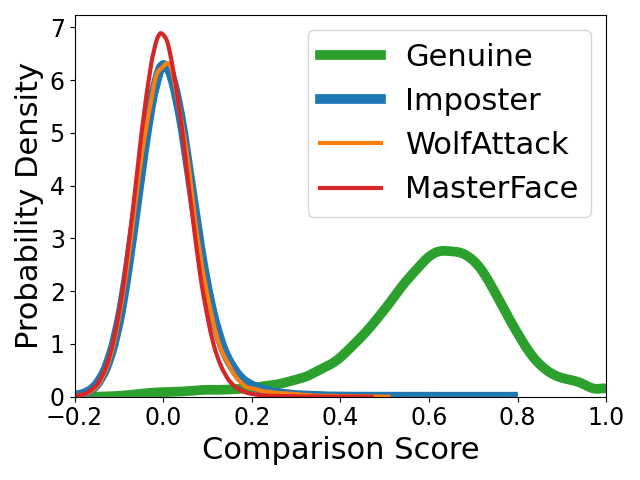}} 
\subfloat[Adience\protect\\Elastic-Arc]{%
     \includegraphics[width=0.135\textwidth]{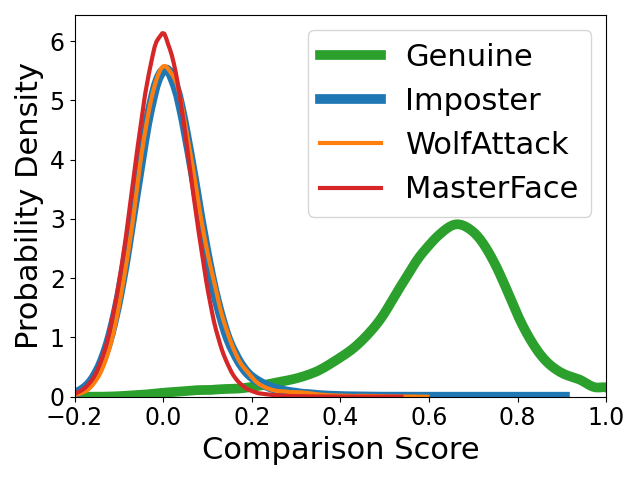}} 
\subfloat[Adience\protect\\Elastic-Cos]{%
     \includegraphics[width=0.135\textwidth]{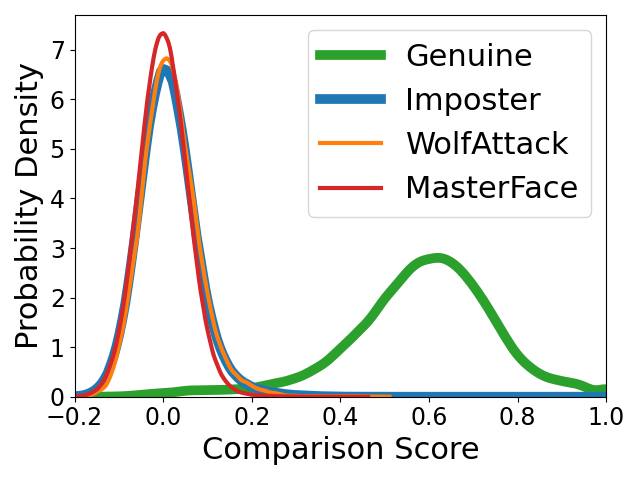}} 
\subfloat[Adience\protect\\MagFace]{%
     \includegraphics[width=0.135\textwidth]{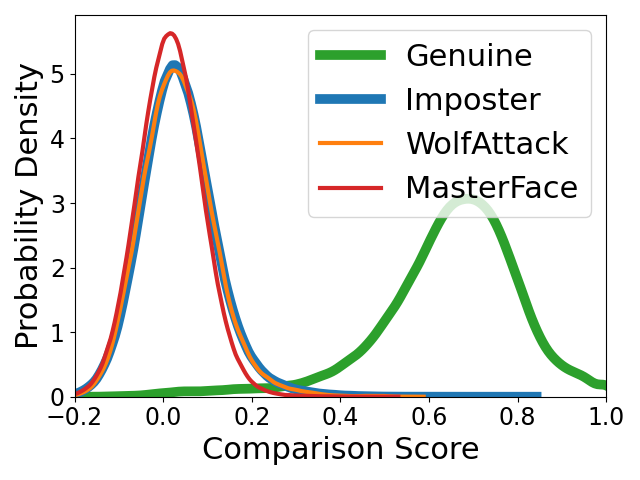}} 
\subfloat[Adience\protect\\QMagFace]{%
     \includegraphics[width=0.135\textwidth]{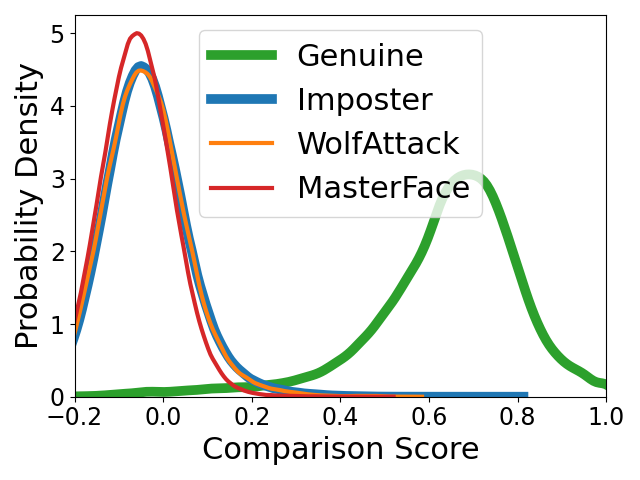}} 
     
\subfloat[LFW\protect\\ArcFace]{%
     \includegraphics[width=0.135\textwidth]{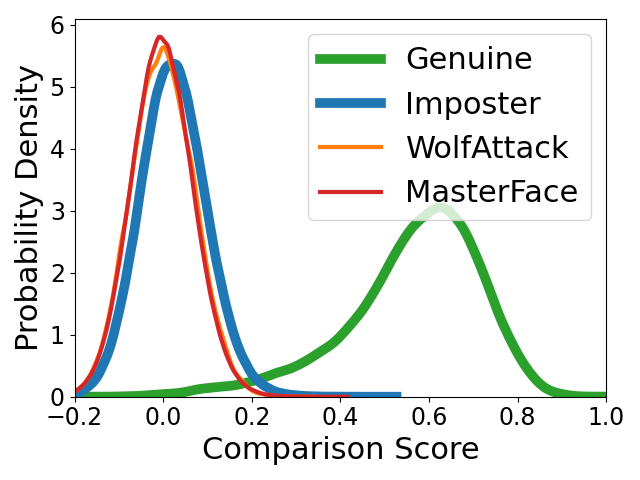}} 
\subfloat[LFW\protect\\CurricularFace]{%
     \includegraphics[width=0.135\textwidth]{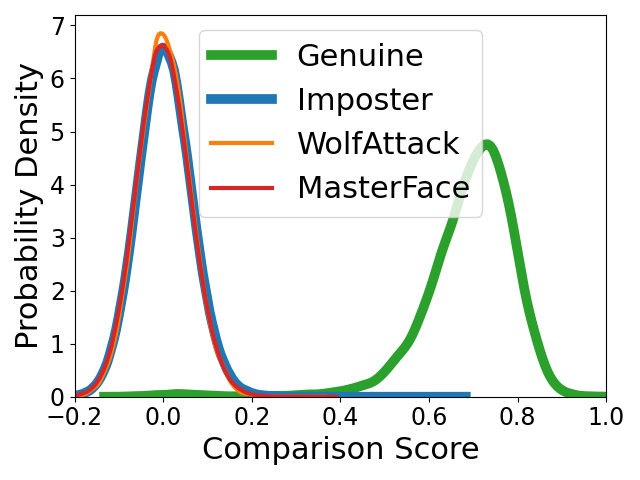}} 
\subfloat[LFW\protect\\Elastic-Arc]{%
     \includegraphics[width=0.135\textwidth]{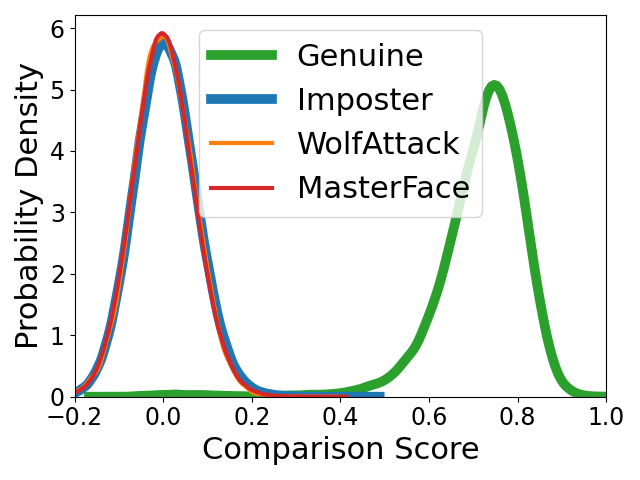}} 
\subfloat[LFW\protect\\Elastic-Cos]{%
     \includegraphics[width=0.135\textwidth]{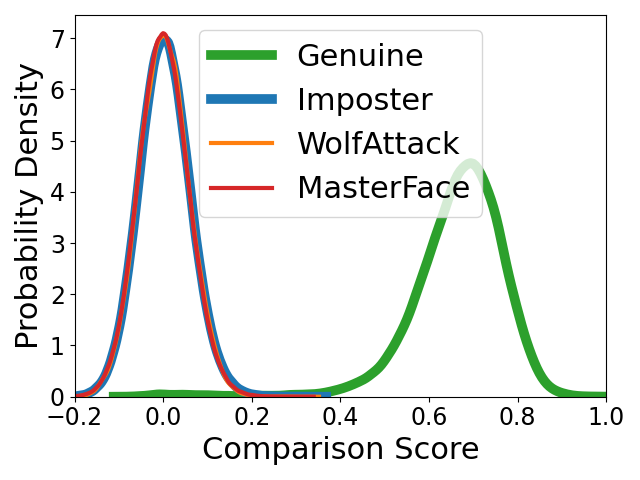}} 
\subfloat[LFW\protect\\MagFace]{%
     \includegraphics[width=0.135\textwidth]{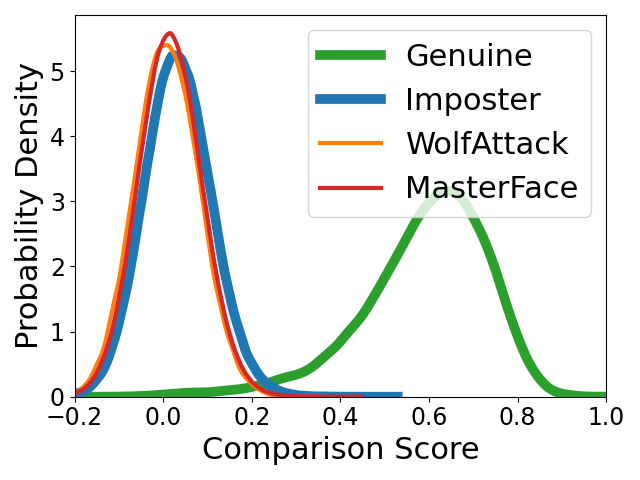}} 
\subfloat[LFW\protect\\QMagFace]{%
     \includegraphics[width=0.135\textwidth]{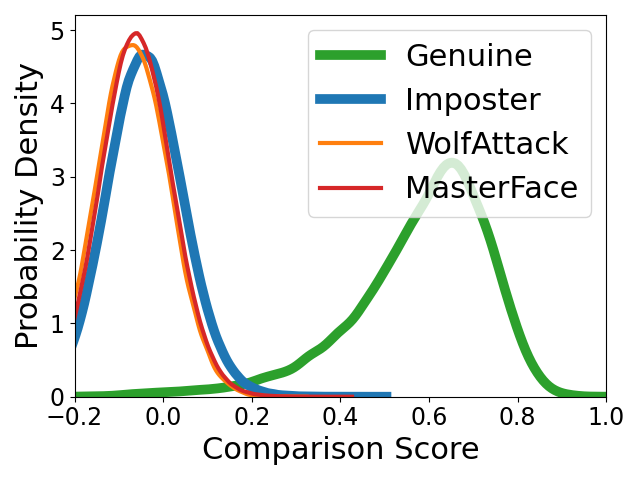}}
     
\subfloat[ColorFeret\protect\\ArcFace]{%
     \includegraphics[width=0.135\textwidth]{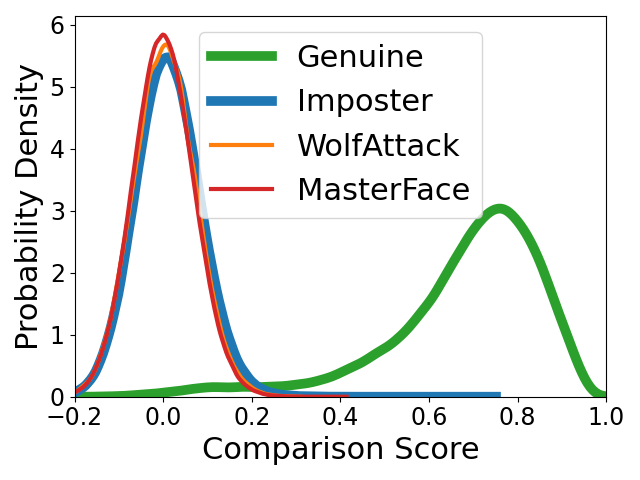}} 
\subfloat[ColorFeret\protect\\CurricularFace]{%
     \includegraphics[width=0.135\textwidth]{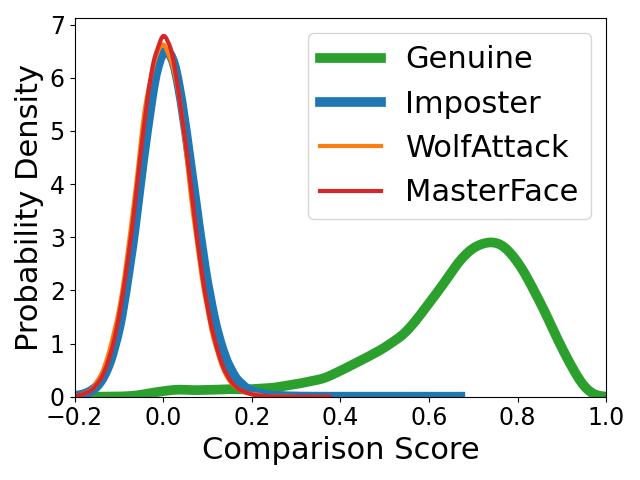}} 
\subfloat[ColorFeret\protect\\Elastic-Arc]{%
     \includegraphics[width=0.135\textwidth]{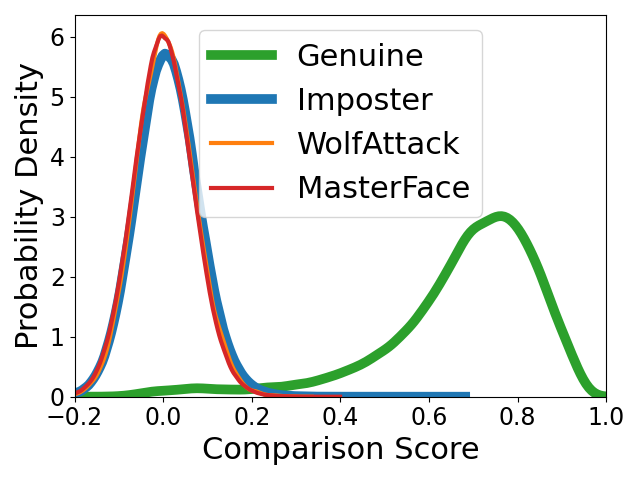}} 
\subfloat[ColorFeret\protect\\Elastic-Cos]{%
     \includegraphics[width=0.135\textwidth]{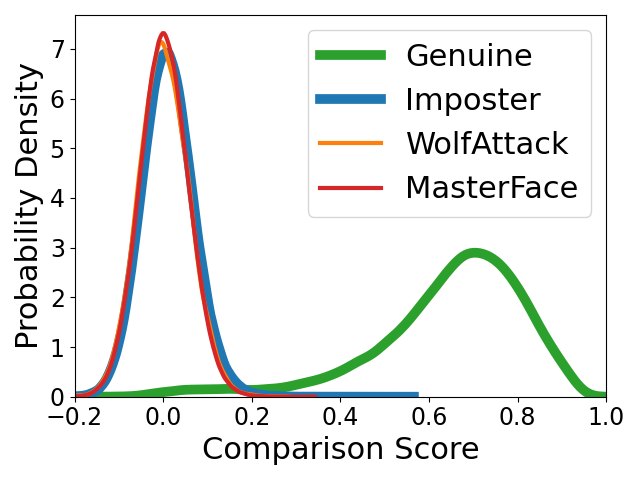}} 
\subfloat[ColorFeret\protect\\MagFace]{%
     \includegraphics[width=0.135\textwidth]{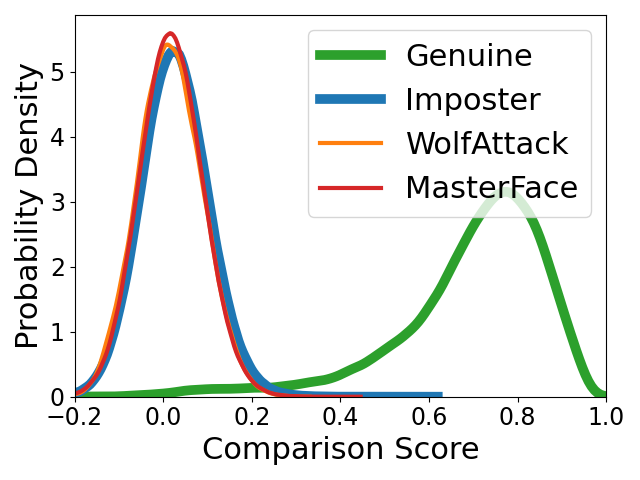}} 
\subfloat[ColorFeret\protect\\QMagFace]{%
     \includegraphics[width=0.135\textwidth]{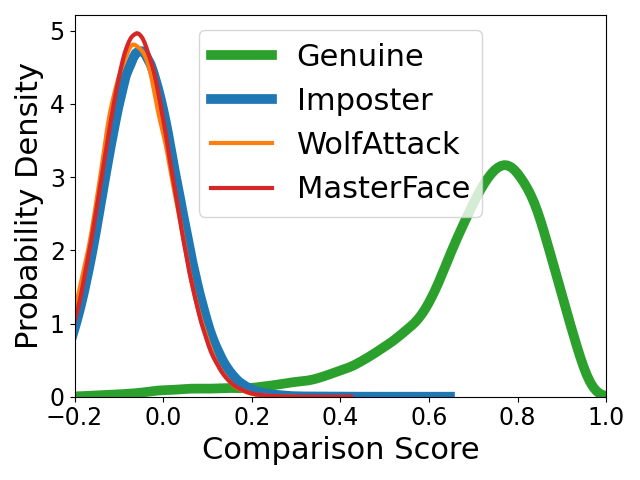}}    
   
\caption{Analysis of the score distributions - The score distributions for the genuine (green) and imposter (blue) comparisons are shown together with the scores distributions that result when comparing the WolfAttack (orange) and MasterFace (red) images against the database. The results are shown for six state-of-the-art FRS on three databases. In all cases the MasterFace attacks closely resemble the imposter distributions showing that the utilized FR models are not vulnerable to these attacks.}
\label{fig:ScoreDistributions}
\end{figure*}

\paragraph{Investigating Score Distributions:}
To get an understanding of how much the MasterFace attacks differ from zero-effort imposter comparisons, Figure \ref{fig:ScoreDistributions} show score distributions for the six FR models on the three databases.
In each plot, the score distributions for genuine (green) and imposter (blue) comparisons are shown together with the score distributions that result by comparing the MasterFace (red) and WolfAttack (orange) images with all images of the database.
The well-separation of the genuine and imposter distributions demonstrates that the utilized FR models perform well.
However, in all cases, the MasterFace attack distributions closely resemble the imposter distributions.
This demonstrates that the utilized FR solutions show no vulnerability to the MasterFace attacks.
This holds also true for the LFW dataset. 
Despite that both methods use large parts of LFW for training, this supports our claim that the success of current MasterFace attacks is based on optimizing the attacks on a specific FR model.





\begin{table*}[]
\renewcommand{\arraystretch}{0.9}
\centering
\caption{Analysis of coverage - The effectiveness of the MasterFace attacks from \cite{DBLP:conf/fgr/ShmelkinWF21} (MasterFace) and \cite{DBLP:conf/icb/NguyenYEM20} (WolfAttack) are analysed. Sample- and identity-coverage rates are reported on three datasets on six state-of-the-art FR models for several decision thresholds ranging from $10^{-1}$ to $10^{-4}$ FMRs. The results show no significant vulnerabilities to MasterFace attacks.}
\label{tab:EmpiricalResults} 
\begin{tabular}{lllrrrrrrrrrrr}
\Xhline{2\arrayrulewidth}
           &            &           &     & \multicolumn{4}{c}{Sample-Coverage (\%) at   FMR}                &                                 & \multicolumn{4}{c}{Identity-Coverage (\%) at FMR}                                               \\
           \cmidrule(rl){5-8} \cmidrule(rl){10-13}
Dataset    & Attack     & FRS        &    & $10^{-1}$ & $10^{-2}$ & $10^{-3}$ & $10^{-4}$ & & $10^{-1}$ & $10^{-2}$ & $10^{-3}$ & $10^{-4}$ \\
\hline
Adience    & WolfAttack & ArcFace        &  & 7.13 & 0.76 & 0.07 & 0.01 &  & 6.74  & 0.50 & 0.05 & 0.01 \\
           &            & CurricularFace &  & 5.93 & 0.72 & 0.14 & 0.03 &  & 6.83  & 0.56 & 0.06 & 0.01 \\
           &            & Elastic-Arc    &  & 6.95 & 0.76 & 0.15 & 0.04 &  & 6.60  & 0.56 & 0.08 & 0.03 \\
           &            & Elastic-Cos    &  & 6.89 & 0.59 & 0.09 & 0.03 &  & 6.96  & 0.47 & 0.05 & 0.01 \\
           &            & MagFace        &  & 6.17 & 0.61 & 0.08 & 0.01 &  & 5.87  & 0.43 & 0.04 & 0.01 \\
           &            & QMagFace       &  & 6.14 & 0.59 & 0.08 & 0.01 &  & 5.85  & 0.43 & 0.04 & 0.01 \\
           \cmidrule(rl){2-13}
           & MasterFace & ArcFace        &  & 4.46 & 0.23 & 0.17 & 0.00 &  & 1.09  & 0.19 & 0.23 & 0.02 \\
           &            & CurricularFace &  & 5.31 & 0.21 & 0.21 & 0.01 &  & 1.11  & 0.21 & 0.25 & 0.04 \\
           &            & Elastic-Arc    &  & 4.83 & 0.20 & 0.21 & 0.01 &  & 1.01  & 0.19 & 0.31 & 0.02 \\
           &            & Elastic-Cos    &  & 4.91 & 0.20 & 0.12 & 0.00 &  & 1.09  & 0.19 & 0.21 & 0.03 \\
           &            & MagFace        &  & 4.38 & 0.19 & 0.18 & 0.00 &  & 0.96  & 0.18 & 0.22 & 0.01 \\
           &            & QMagFace       &  & 4.35 & 0.19 & 0.18 & 0.00 &  & 0.96  & 0.18 & 0.22 & 0.02 \\
           \hline
LFW        & WolfAttack & ArcFace        &  & 2.96 & 0.10 & 0.00 & 0.00 &  & 3.59  & 0.20 & 0.01 & 0.00 \\
           &            & CurricularFace &  & 5.59 & 0.34 & 0.02 & 0.00 &  & 5.73  & 0.60 & 0.05 & 0.01 \\
           &            & Elastic-Arc    &  & 5.59 & 0.36 & 0.02 & 0.00 &  & 5.40  & 0.53 & 0.04 & 0.01 \\
           &            & Elastic-Cos    &  & 6.11 & 0.39 & 0.02 & 0.00 &  & 5.78  & 0.59 & 0.05 & 0.01 \\
           &            & MagFace        &  & 3.07 & 0.13 & 0.01 & 0.00 &  & 3.65  & 0.22 & 0.02 & 0.00 \\
           &            & QMagFace       &  & 3.08 & 0.13 & 0.01 & 0.00 &  & 3.65  & 0.22 & 0.02 & 0.00 \\
           \cmidrule(rl){2-13}
           & MasterFace & ArcFace        &  & 4.20 & 0.22 & 0.02 & 0.01 &  & 0.87  & 0.21 & 0.03 & 0.02 \\
           &            & CurricularFace &  & 8.73 & 0.73 & 0.06 & 0.02 &  & 1.05  & 0.48 & 0.07 & 0.02 \\
           &            & Elastic-Arc    &  & 8.53 & 0.71 & 0.06 & 0.02 &  & 1.04  & 0.47 & 0.07 & 0.02 \\
           &            & Elastic-Cos    &  & 8.94 & 0.73 & 0.06 & 0.01 &  & 1.03  & 0.46 & 0.07 & 0.02 \\
           &            & MagFace        &  & 5.53 & 0.33 & 0.03 & 0.01 &  & 0.89  & 0.28 & 0.05 & 0.02 \\
           &            & QMagFace       &  & 5.53 & 0.34 & 0.03 & 0.02 &  & 0.89  & 0.29 & 0.04 & 0.02 \\
           \hline
ColorFeret & WolfAttack & ArcFace        &  & 4.76 & 0.32 & 0.01 & 0.00 &  & 10.21 & 1.58 & 0.13 & 0.00 \\
           &            & CurricularFace &  & 4.40 & 0.29 & 0.01 & 0.00 &  & 10.80 & 1.49 & 0.13 & 0.00 \\
           &            & Elastic-Arc    &  & 5.06 & 0.33 & 0.02 & 0.00 &  & 10.18 & 1.49 & 0.10 & 0.00 \\
           &            & Elastic-Cos    &  & 4.81 & 0.29 & 0.02 & 0.00 &  & 10.65 & 1.51 & 0.12 & 0.00 \\
           &            & MagFace        &  & 4.72 & 0.26 & 0.00 & 0.00 &  & 10.24 & 1.35 & 0.02 & 0.00 \\
           &            & QMagFace       &  & 4.74 & 0.26 & 0.00 & 0.00 &  & 10.26 & 1.37 & 0.02 & 0.00 \\
           \cmidrule(rl){2-13}
           & MasterFace & ArcFace        &  & 6.50 & 0.40 & 0.03 & 0.01 &  & 3.45  & 0.75 & 0.18 & 0.08 \\
           &            & CurricularFace &  & 7.02 & 0.43 & 0.04 & 0.01 &  & 3.33  & 0.76 & 0.19 & 0.07 \\
           &            & Elastic-Arc    &  & 6.96 & 0.45 & 0.04 & 0.01 &  & 3.85  & 0.77 & 0.20 & 0.05 \\
           &            & Elastic-Cos    &  & 6.80 & 0.38 & 0.03 & 0.01 &  & 2.94  & 0.73 & 0.16 & 0.05 \\
           &            & MagFace        &  & 7.09 & 0.46 & 0.04 & 0.01 &  & 3.33  & 0.73 & 0.19 & 0.12 \\
           &            & QMagFace       &  & 7.11 & 0.46 & 0.04 & 0.01 &  & 3.33  & 0.73 & 0.19 & 0.09 \\
           \Xhline{2\arrayrulewidth}              
\end{tabular}
\end{table*}

\subsection{Theoretical Analysis}
\label{sec:TheoreticalAnalysis}

The results from the empirical analysis suggests that the MasterFace generation exploits regions in the embedding space of a specific FR model with a weak identity separation.
Since the generalization capabilities of FR models increases drastically with every year, regions that MasterFace attacks might exploit continuously decrease. 
In this section, we analyse the border case of an FR model with perfect identity-separation as described in Section \ref{sec:Methodology}.
 
\begin{figure*}
\centering
\subfloat[Face Capacity \label{fig:Capacity}]{%
     \includegraphics[width=0.27\textwidth]{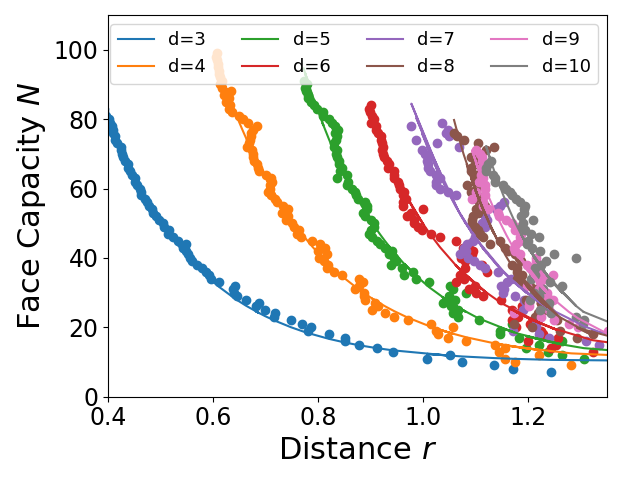}} 
\subfloat[Maximum MasterFace Coverage  \label{fig:Coverage}]{%
     \includegraphics[width=0.27\textwidth]{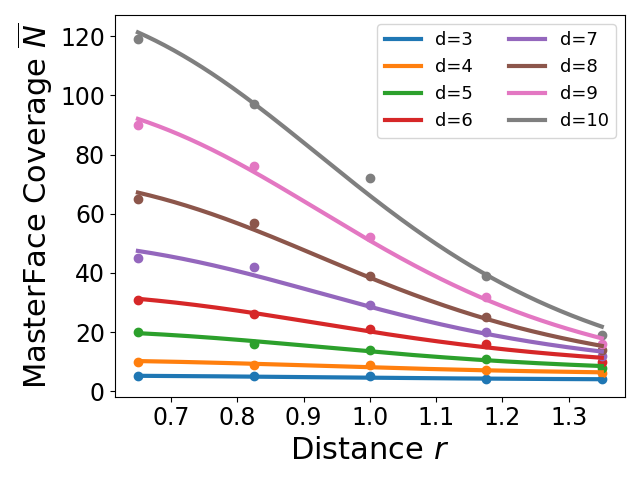}}
\subfloat[High-Dimensional Behavior \label{fig:Behaviour}]{%
     \includegraphics[width=0.274\textwidth]{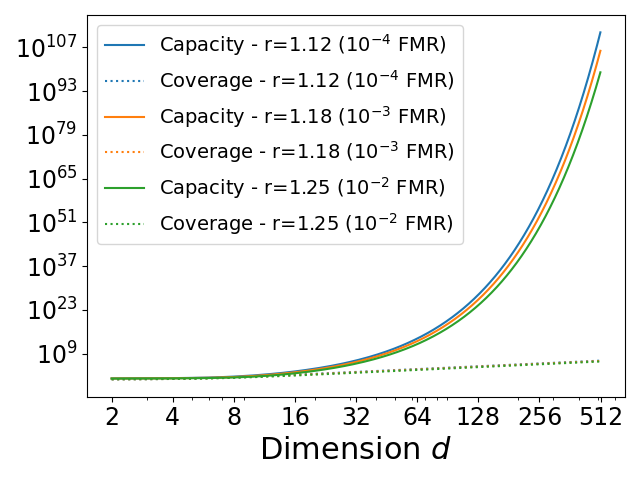}}
\caption{Analysis of the face capacity and MasterFace coverage fits - In (a), the maximum number of identities $N$ with a distance of greater than $r$ to each other are shown (capacity). In (b), the same is represented for the maximum number of identities that have at distance of greater $r$ to each other while the respective distance to a potential MasterFace is $r$ (maximum coverage). The dots represent the solutions of their optimization problems, while the solid lines represent the fits. In (c), the high-dimensional behaviour of both functions are shown.} 
\label{fig:TheoreticalAnalysis_Fits}
\end{figure*}

Figure \ref{fig:TheoreticalAnalysis_Fits} analyses the face capacity and MasterFace coverage fits defined in Equations \ref{eq:FitCapacity} and \ref{eq:FitCoverage}.
In Figures \ref{fig:Capacity} and \ref{fig:Coverage}, the solutions of the optimization problem are shown as dots, while the solid lines represent the fits.
In Figure \ref{fig:Capacity}, an exponential behaviour concerning $r$ is observed, while Figure \ref{fig:Coverage} follows a sigmoid function with respect to $r$.
Both figures demonstrate that the functions are well suitable for the fit and result in reasonable approximations.
Figure \ref{fig:Behaviour} extrapolates\footnote{Please note that these functions result from an extrapolation process and thus, might not show the optimal values.
We recommend to use these only to get a general idea of the magnitude of the face capacity and the maximum MasterFace coverage.} the face capacity and the maximum MasterFace coverage to higher dimensions that are more frequently used in FR models, such as $d=128$ and $d=512$.
The coverage and capacity functions are shown for three different $r$ values that refer to different FMR thresholds of the utilized models ($r=1.12$ for $10^{-4}$ FMR, $r=1.18$ for $10^{-3}$ FMR, $r=1.25$ for $10^{-2}$ FMR). 
Independent of the choice of $r$, the face capacity shows a stronger (exponential) growth than the (polynomial growth) of the maximum MasterFace coverage for higher dimensions.

\begin{figure}
\centering
\includegraphics[width=0.4\textwidth]{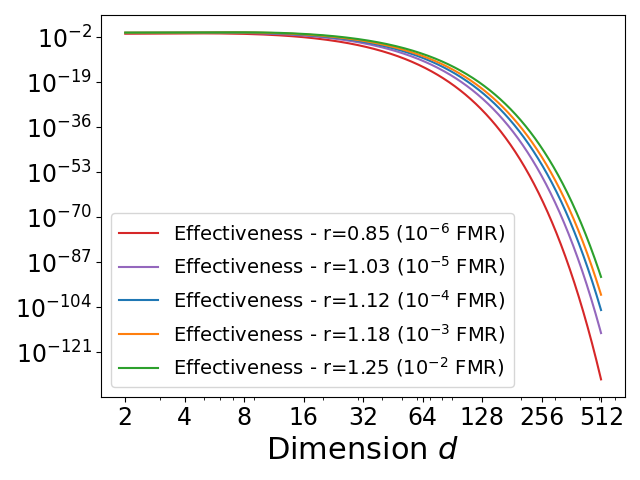}
\caption{Analysis of the maximum MasterFace effectiveness $\eta$ - The maximal proportion of all perfectly-separable identities in the embeddings space that can successfully match with a MasterFace are shown for different $d$ and $r$. For embedding dimensions usually used in FRS ($d\geq 128)$), the effectiveness becomes very small. This demonstrates that FRS with a perfect identity separation are not significantly vulnerable to MasterFace attacks.} 
\label{fig:MasterFaceEffectiveness}
\end{figure}

This has a significant impact on the maximum MasterFace effectiveness as shown in Figure \ref{fig:MasterFaceEffectiveness}.
With higher dimensions $d$ (and lower decision thresholds $r$) the maximum MasterFace effectiveness $\eta$ drops significantly.
Consequently, effective MasterFace attacks are only possible in very low-dimensional face representations (e.g. $d<10$) assuming a perfect identity-separation in the face space.

\section{Summary of Observations}

%
%
%

In contrast to previous works, the experiments of this work make use of six state-of-the-art FR model, multiple testing datasets of significantly higher size and variance, and cross-dataset and cross-model evaluations.
This leads to some key observations.
Based on the empirical and theoretical investigations of this work, we observe
\begin{enumerate}
\item \textbf{Low sample- and identity coverage rates of MasterFace attacks in cross-model evaluations} - While in \cite{DBLP:conf/fgr/ShmelkinWF21} identity-coverage between 17-22\% at an FMR threshold of $10^{-3}$ were reported, our experiments show significant lower identity-coverage between 0.01-0.31\% at the same threshold when the attacks are not trained on the specific FR model.
\item \textbf{Score distributions of MasterFace attacks very similar to imposter distributions} - For all FR model and dataset combination, we observe that the score distribution of MasterFace attacks closely resemble the distribution of imposter comparisons in cross-dataset and cross-model evaluation settings.
\item \textbf{No performance improvement of MasterFace attacks trained on testing data} - Despite that both MasterFace attack methods were trained on large parts of the LFW dataset, we observe no improvement of the MasterFace attack performance on this dataset compared to the others.
\item \textbf{Insignificant maximum MasterFace effectiveness in face spaces with perfect identity-separation} - The theoretical analysis showed that the maximum number of perfectly-separable identities in a face space (capacity) is several magnitudes higher  than the number of perfectly-separable identities that a MasterFaster can maximally cover. 
\end{enumerate}

Observation 1 and 2 demonstrates a low generalization of MasterFace attacks to cross-modal and cross-dataset evaluation settings and shows that these MasterFace images perform similar to zero-effort imposter comparisons.
Observation 3 eliminates the cross-dataset factor since training MasterFace attacks on part of the training data does not increase the success of these attacks.
\textit{Consequently, we assume that an FR model might be only vulnerable to MasterFace attacks when these are specifically trained on the individual model.}
Observation 4 states that for a face space with perfect identity-separation, even an optimal MasterFace will not cover a significant portion of possible identities.
With the current trend of increasing generalizability in FRS, \textit{this indicates that the vulnerability of future FRS to MasterFace attacks might further decrease.}

\section{Conclusion}










In this work, we comprehensively analysed the generalizability of MasterFace attacks empirically and theoretically.
Contrarily to previous works, our empirical investigations included the use of six state-of-the-art FR model, larger testing datasets, and cross-dataset and cross-model evaluation settings.
The results show a MasterFace attack performance, similar to zero-effort imposter, in our evaluation setting.
This indicates a low cross-model generalizability, meaning that MasterFace attacks might only be successful when the same FRS is used for training and testing.
In the theoretical investigation, we defined and investigated the face capacity and the maximum MasterFace coverage to analyse the maximum MasterFace effectiveness under the assumption of an FRS with perfect identity-separation.
The investigations showed an insignificantly low maximum MasterFace effectiveness for such FRS.
Since research on FR still aims for higher generalizability, this indicates that the vulnerability of future FRS for MasterFaces might further decrease.
We emphasize that future works might investigate the use of MasterFaces for analysing and enhancing the robustness of FR models.

\paragraph{Acknowledgement}
This work was carried out during the tenure of an ERCIM ’Alain Bensoussan‘ Fellowship Programme.
Parts of this research work has been funded by the German Federal Ministry of Education and Research and the Hessen State Ministry for Higher Education, Research and the Arts within their joint support of the National Research Center for Applied Cybersecurity ATHENE.
Portions of the research in this paper use the FERET database of facial images collected under the FERET program, sponsored by the DOD Counterdrug Technology Development Program Office.

{\small
\bibliographystyle{ieee}
\bibliography{egbib}
}

\newpage
\clearpage

\section*{Supplementary}

Due to the strict page limit, the paper focuses on only briefly describing the functions for the face capacity (Eq. \ref{eq:FitCapacity}) and the maximum MasterFace coverage (Eq. \ref{eq:FitCoverage}) in Section \ref{sec:MaximumMFEffectiveness} and discussing its suitability in Section \ref{sec:MaximumMFEffectiveness}.
In this section, we will explain the choice of these functions and their suitability in more details.

\begin{figure*}[bh]
\centering
\subfloat[Face Capacity Para. $A$ \label{fig:CapaPara_A}]{%
     \includegraphics[width=0.25\textwidth]{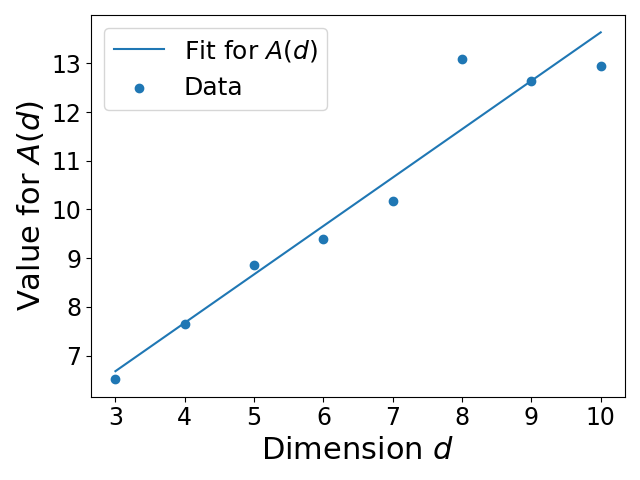}} 
\subfloat[Face Capacity Para. $B$ \label{fig:CapaPara_B}]{%
     \includegraphics[width=0.25\textwidth]{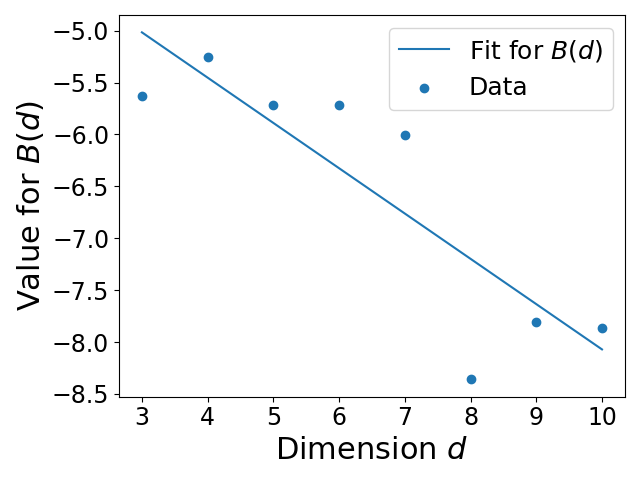}} 
\subfloat[MasterFace Coverage Para. $\overline{A}$ \label{fig:CovPara_A}]{%
     \includegraphics[width=0.25\textwidth]{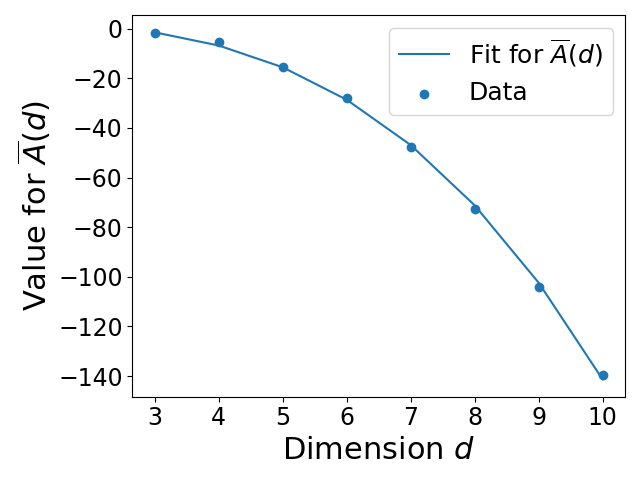}}  
\subfloat[MasterFace Coverage Para. $\overline{B}$ \label{fig:CovPara_B}]{%
     \includegraphics[width=0.25\textwidth]{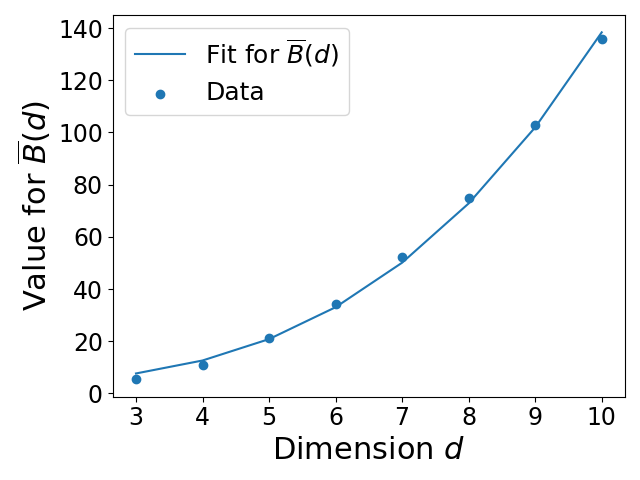}} 
\caption{Parameter values for the face capacity and MasterFace coverage for single dimensions $d$.}
\label{fig:FittingDetails}    
\end{figure*}

\subsection*{Deriving the Functions}
For both, the face capacity and the maximum MasterFace coverage, we computed the optimal solutions for $d=[3,\dots,10]$ with a Sequential Least Squares Programming (SLSQP) algorithm as described in Section \ref{sec:MaximumMFEffectiveness}. 
These solutions are shown as dots in Figures \ref{fig:Capacity} and \ref{fig:Coverage}.
For finding the best functions that describe these solutions, we tried a wide range of  functions such as different polynomial, exponential, and sigmoid functions and evaluated these on single fixed dimensions $d$.
For the final function, we choose the simplest function that described the behaviour of the data well.
This resulted in a face capacity function
\begin{align}
N(r,d) = \exp [A(d) + r \, B(d)] \label{eq:Sup_capa}
\end{align}
with two parameters $A$ and $B$ describing a linear function regarding $r$ in the argument of the exponential function.
Each $d$ resulted in different parameter values for $A$ and $B$.
These values are shown in Figure \ref{fig:CapaPara_A} and \ref{fig:CapaPara_B} for the different dimensions $d$.
Based on the linearity of these, we fit the following linear functions
\begin{align}
A(d) &= \alpha \, d + \beta \\
B(d) &= \gamma \, d + \delta
\end{align}
on the parameter points, resulting in the parameter values $\alpha, \beta, \gamma, \delta$ as shown in Table \ref{tab:LearnedParameters}.
Inserting $A(d)$ and $B(d)$ into Eq. \ref{eq:Sup_capa} leads to the used face capacity function in Eq. \ref{eq:FitCapacity}.

The same procedure was applied for the maximum MasterFace coverage.
Similarly, to the face capacity function, the function for the maximum MasterFace coverage
\begin{align}
\overline{N}(r,d) = \overline{A}(d) \cdot \sigma\left( \phi \, (   r^{\epsilon}-1) \right) + \overline{B}(d), \label{eq:Sup_cov}
\end{align}
consists of two parameters $\overline{A}$ and $\overline{B}$ that are determined for each $d$ separately.
However, in contrast the face capacity, the MasterFace coverage follows a sigmoid shape with respect to $r$ (see Figure \ref{fig:Coverage}).
Consequently, we used a modified sigmoid function that is squeezed and shifted to the desired shape.
The values of $\overline{A}$ and $\overline{B}$ for different dimensions $d$ are shown in Figures \ref{fig:CovPara_A} and \ref{fig:CovPara_B}.
These follow a behaviour that can be well described through polynomial functions of third order 
\begin{align}
\overline{A}(d) = \overline{\alpha} \, d^3 + \overline{\beta} \\
\overline{B}(d) = \overline{\gamma} \, d^3 + \overline{\delta}.
\end{align}
As before, the resulting parameters are shown in Table \ref{tab:LearnedParameters} and inserting $\overline{A}(d)$ and $\overline{B}(d)$ into Eq. \ref{eq:Sup_cov} results in the used maximum MasterFace coverage function \ref{eq:FitCoverage}.

\subsection*{Function Suitability}

The advantage of dividing the fitting process into two steps is that we can directly observe the quality of the fits with respect to the dimension parameter $d$ as shown in Figure \ref{fig:FittingDetails}.
For the face capacity function, we can observe an outlier for $d=8$ that slightly degrade the quality of the fit.
However, assuming a linear function for $A(d)$ seems fitting (see Figure \ref{fig:CapaPara_A}) and will probably result in reasonable parameter values for larger $d$.
For the function $B(d)$, assuming a linear function was more of safe choice since the stability of the solver and the outlier for $d=8$ results in a less clear function behaviour.
Combing the observations for $A(d)$ and $B(d)$, the face capacity function $N(d,r)$ might significantly differ from the optimal values.
Consequently, we recommend to use this function only to get a general idea of the magnitude of the face capacity as we mentioned and did in the main paper.
For the maximum MasterFace coverage, both functions $\overline{A}$ and $\overline{B}$ well represents the parameter values as observable in Figures \ref{fig:CovPara_A} and \ref{fig:CovPara_B}.
This indicates that the maximum MasterFace coverage function $\overline{N}$ might result in highly suitable estimates with respect to $d$.

\end{document}